\pdfoutput=1

\documentclass[11pt]{article}

\usepackage{ACL2023}

\usepackage{times}
\usepackage{latexsym}

\usepackage{amssymb}
\usepackage{amsmath}
\usepackage{graphicx}
\usepackage{hyperref}

\usepackage[T1]{fontenc}

\usepackage[utf8]{inputenc}

\usepackage{microtype}

\usepackage{inconsolata}

\title{Memory-efficient Stochastic methods for Memory-based Transformers}

\author{Vishwajit Kumar Vishnu \\
  Indian Institute of Technology Madras \\
  \texttt{cs18s039@cse.iitm.ac.in} \\\And
  C. Chandra Sekhar \\
  Indian Institute of Technology Madras \\
  \texttt{chandra@cse.iitm.ac.in} \\}

\begin{document}
\maketitle
\begin{abstract}
Training Memory-based transformers can require a large amount of memory and can be quite inefficient. We propose a novel two-phase training mechanism and a novel regularization technique to improve the training efficiency of memory-based transformers, which are often used for long-range context problems. For our experiments, we consider transformer-XL as our baseline model which is one of memory-based transformer models. We show that our resultant model, Skip Cross-head Transformer-XL, outperforms the baseline on character level language modeling task with similar parameters and outperforms the baseline on word level language modelling task with almost 20\% fewer parameters. Our proposed methods do not require any additional memory. We also demonstrate the effectiveness of our regularization mechanism on BERT which shows similar performance with reduction in standard deviation of scores of around 30\% on multiple GLUE tasks. \footnote{ The implementation details are added in Appendix and   the code is available at following $\href{https://github.com/vishwajit-vishnu/Memory-efficient-Stochastic-methods-for-Memory-based-Transformers}{github\_link}$.}

\end{abstract}

\section{Introduction}

Transformers with memory component are used for long range dependency problems in the area of language modelling, speech technology and reinforcement learning \citep{DBLP:conf/iclr/RaePJHL20}. But, training these models efficiently and effectively requires a large amount of memory. In this work, we show two stochastic methods to improve the performance of these models without requirement of any extra memory. 

In this work, we use language modelling as our primary task to conduct experiments and ablation studies. Language modelling is a long range dependency problem of predicting next token given a sequence of previous tokens. Language models estimate joint probability $P(\mathbf{x})$  for a token sequence $\mathbf{x}= (x_1, x_2, .. , x_T) $.

\[
    P(\mathbf{x}) = \prod_{t=2}^{T} P(x_t | x_{t-1}, x_{t-2}, .. , x_1)
\]

We use Transformer-XL, a memory-based transformer model as baseline model. We apply our proposed methods on our baseline and introduce the resultant model as Skip Cross-head Transformer-XL. Refer Appendix \ref{appendix:recap_txl} for a recap on transformer-XL.

Following are some of the highlights of our work:

\begin{itemize}
    \item Our novel two phase Skip-Retain training mechanism attends to a longer context without requirement of any additional memory and helps in improving the deep representations. This method can be applied to memory-based models. Our proposed model outperforms the baseline with 20\% fewer parameters (Table \ref{table:wt103}).
    \item  Our novel Stochastic cross-head attention mechanism leverages its regularization effect with query-key interactions among different heads of same layer. This helps in attending to the information from different subspaces and acts as  an internal regularizer. Also, unlike previous methods like \citep{DBLP:journals/corr/abs-2003-02436}, this mechanism  does not require any additional parameters. This can be used as plug and play for other transformer based models as well. We conducted additional experiments on GLUE tasks \citep{DBLP:conf/iclr/WangSMHLB19} with BERT \citep{DBLP:conf/naacl/DevlinCLT19} (Table \ref{table:bert}) to show the effectiveness of this mechanism. This mechanism improves the standard deviation of the results by almost 30\% for multiple GLUE tasks. In pruning of individual heads experiment (similar to \citet{DBLP:conf/nips/MichelLN19}), it shows a reduction in standard deviation of 50\% in all layers, showing the redistribution of information among heads without any reduction in the performance.
    \item Another advantage of these methods is that these methods are employed only at the time of training or fine-tuning and thus, these methods do not affect the inference time of the models.
\end{itemize}

\section{Proposed Methods}

\subsection{Skip-Retain Training}

We propose a novel two phase training mechanism that can attend to longer context without requirement of any additional memory.

\subsubsection{Training Mechanism}

\begin{figure}
\hspace{\fill}
\centering
  \includegraphics[scale=0.127]{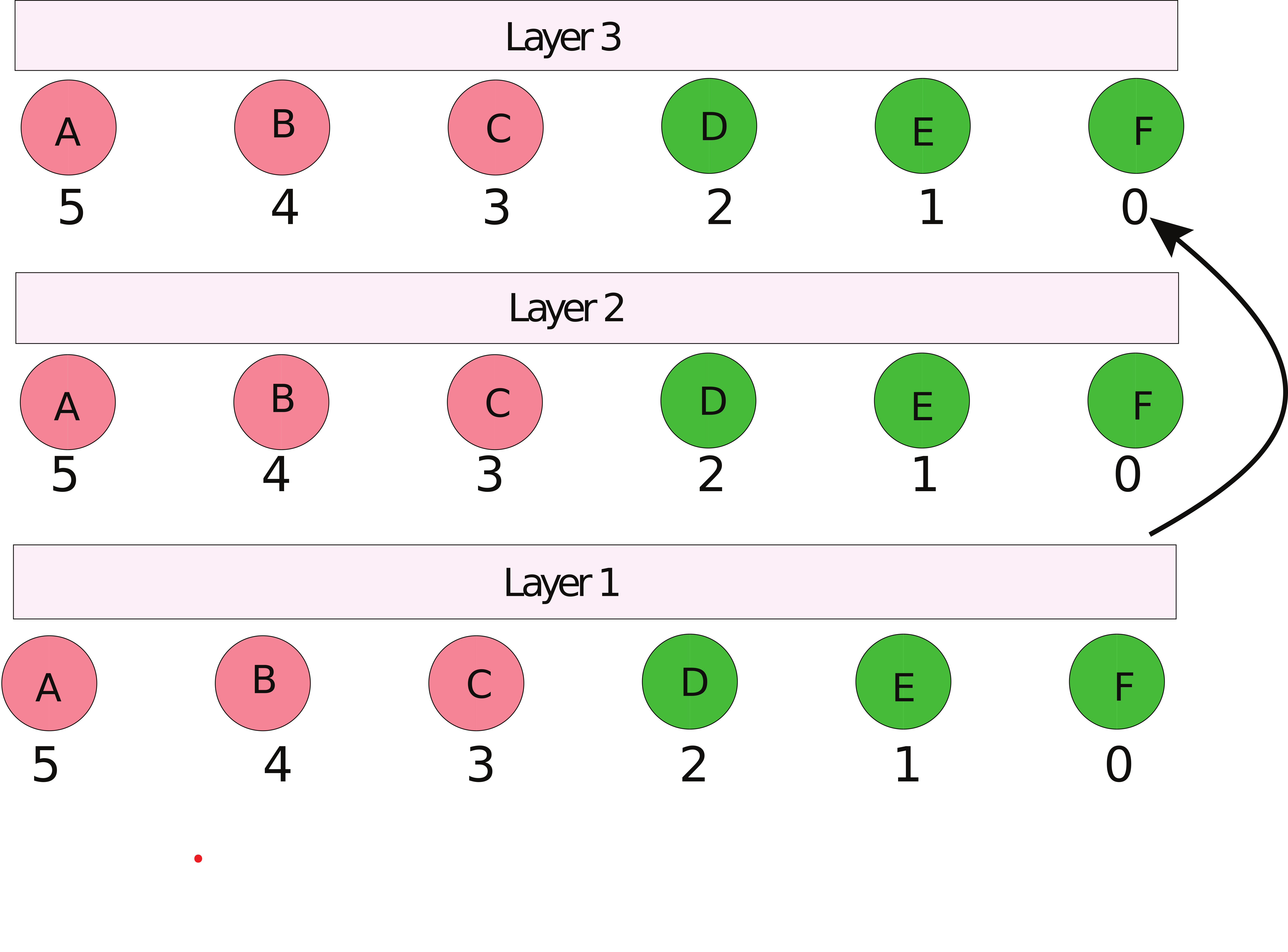}
      \caption{An example of Phase 1 for the three layers of a Transformer-XL. Light red colour denotes the past activations stored in the memory component (memory length, $M=3$). Green colour denotes the current input tokens (block size, $L=3$). In this example, we are giving enlish alphabet sequence A, B, C, D, E,..., Z as input. }

    \label{fig:fig1}
\end{figure}

\begin{figure}
\hspace{\fill}
\centering
\includegraphics[scale=0.127]{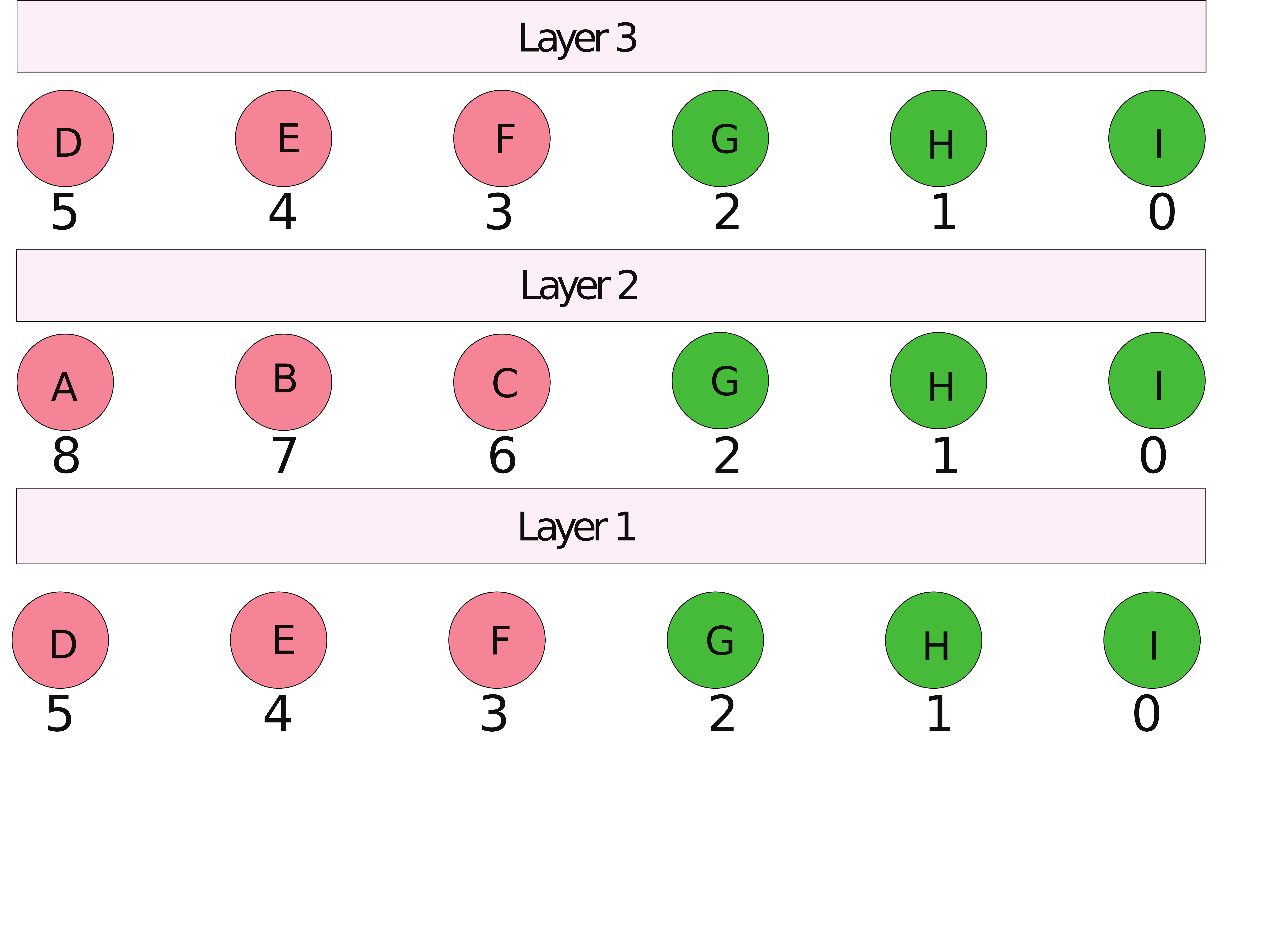}
  \caption{The relative positions after skipping the middle layer in Figure \ref{fig:fig1}. Due to skipping of the middle layer, the activations stored in memory component are not updated. In this way, with the same size memory component, the token "I" can attend to token "A" in middle layer whereas other layers cannot do so, thereby it can attend to a longer context.}
    \label{fig:fig2}
\end{figure}

A memory-based transformer model stores the previous time step's ($t-1$ step's) input activations in the memory component of each layer. In our proposed method, we skip a layer of the model with some probability. When we skip a layer, we retain the activations in the memory component of the skipped layer. In this way, in the next time step ($(t+1)^{th}$ time step), the memory component of the skipped layer contains the input activations of $(t-1)^{th}$ time step and the non-skipped layers stores the activations of ${t}^{th}$ time step. This has been depicted visually in Figure \ref{fig:fig1} and \ref{fig:fig2} where relative positions of the tokens for $block\_size=3$ is shown (block size is the number of tokens given as input at once in the current time step to the transformer model(refer Appendix \ref{appendix:recap_txl})). The input tokens are shown in green and the most recent token is given a relative position of $0$, i.e. higher the relative position of the token, older the token. This first phase of the training mechanism is called Skip-Retain phase. Thus, it helps in attending to a much larger context without requiring any extra memory. We show the expected context of this method in subsection \ref{subsec:expected_context}.

Since skipping of layers is not continued during evaluation, it creates a distribution shift in the relative positions used during training and evaluation. Thus, after the model's convergence in Skip-Retain phase, we train the model like vanilla training i.e. without skipping of the layers. We summarize our complete training mechanism as follows: \\
\begin{itemize}
    \item (Phase 1) Skip-Retain Phase: We train the model by skipping layers with some probability and the memory component of the skipped layers retain the activations stored i.e. the memory component of skipped layer is not updated.
    \item (Phase 2) Vanilla Training Phase: After convergence of the model in Phase 1, we train the model without any skipping of layers.
\end{itemize}

We conducted experiments with different ways of skipping layers like skipping layers uniformly, skipping as a function of layer and skipping the initial and final layer of the model. The probability of skipping $i^{th}$ layer is denoted as $p_{skip}(i)$. We experimented with different ways of skipping layers which are shown below:

\begin{equation}
\label{eqn:functionalskip}
    p_{skip}(i)= 
    \begin{cases}
        0.5 \frac{(i-1)}{N} , \text{ if  } i \neq N \\
        0 , \text{  otherwise}
    \end{cases}
\end{equation}

\begin{equation}
\label{eqn:uniformskip}
  p_{skip}(i)= p , \forall i \in \{1,2,.. N\}
\end{equation}

\begin{equation}
\label{eqn:uniformskip_initial}
    p_{skip}(i)= 
    \begin{cases}
        0,  \text{if  $i$ $=$ $1$}  \\
        p  \text{ ,  otherwise}
    \end{cases}
\end{equation}

\begin{equation}
\label{eqn:uniformskip_final}
    p_{skip}(i)= 
    \begin{cases}
        0,  \text{if  $i$ $=$ $N$}  \\
        p  \text{ ,  otherwise}
    \end{cases}
\end{equation}

\begin{equation}
\label{eqn:uniformskip_both}
    p_{skip}(i)= 
    \begin{cases}
        0,  \text{if  $i$ $=$ $1$ or $i$ = $N$}  \\
        p  \text{ ,  otherwise}
    \end{cases}
\end{equation}
where $N$ is the number of layers in the model.

\subsubsection{Expected context}
\label{subsec:expected_context}

In our experiments we found out the following equation to be most effective (refer subsection \ref{Ablation_Study} for ablation studies):
\[
    p_{skip}(i)= 
    \begin{cases}
        0.5 \frac{(i-1)}{N} , \text{ if  } i \neq N \\
        0 , \text{  otherwise}
    \end{cases}
\]

During the Skip-Retain phase (Phase 1), the layer i is skipped with above probability $p_{skip}(i)$. Usually the memory component stores a fixed context of  $M$ past tokens. But, when a layer is skipped it stores the context from past $2M$ to $M+1$ past tokens (assuming memory component length $M$ is equal to block size $B$), thereby  stochastically attending context  till $2M$. We denote the maximum previous context that can be attended using this method as $m$. So, the expected context is:

\[
     E(\widetilde{m})  =  \sum_{i=1}^{N} p_{skip}(i).2M
\]
\[
    E(\widetilde{m})  =  2M \sum_{i=1}^{N-1} 0.5 \frac{(i-1)}{N}
\]
\begin{equation}
\label{eqn:expected_context}
     E(\widetilde{m})  \approx  M \frac{(N-3)}{2}
\end{equation}

So, for number of layers, $N>5$ , the method increases the effective context stochastically. For a network with  15 layers can stochastically attend to $M \frac{(15-3)}{2} = 6M$, i.e. 6 times the context with constant computational memory.

\subsection{Stochastic Cross-Head Attention}

We propose a stochastic method for regularization that helps in redistribution of information among heads of same layer in transformer models. This method is shown to address the problem of redundant information among heads in a layer (refer Table \ref{table:prune}). \citet{DBLP:conf/nips/MichelLN19} had shown that
only few heads contain information that significantly reduce performance and most heads contain redundant information in a layer.

In our proposed method, the query of one head attends to key and value representations of some other head in each layer with probability $\beta$. Each transformer layer consists of multiple heads where each head has a query, key and value representations. We denote these heads as $h_1, h_2, ..., h_{N_h}$. When random generated probability $u$ is more than $\beta$, then each head's query attends to same head's key and value representations, i.e., $h_1$'s query attends to $h_1$'s key and value(or sequence ($h1, h_2, ..., h_{N_h}$) query representations match to sequence ($h_1, h_2, ..., h_{N_h}$) key and value representations).
When $u < \beta$ each head's query attends to some random head's key and value representations, i.e., $h_1$'s query attends to some random head's key and value(or sequence ($h1, h_2, ..., h_{N_h}$) query representations match to random\_permutation(sequence ($h_1, h_2, ..., h_{N_h}$)) key and value representations). This mechanism is named as stochastic cross-head attention (SCH attn) as it redistributes information across heads of same layer stochastically. $\beta$ is referred as stochastic cross-head attention probability in following sections. The attention score can be formally put as follows (this can be used for vanilla attention mechanism models like BERT \citep{DBLP:conf/naacl/DevlinCLT19}): 

\begin{equation}
     A_{ij,\textbf{M}\textbf{N}} = x_{i}^T  W_{q \textbf{M}}^T  W_{k \textbf{N}} x_{j}
\end{equation}

where head M and head N will be same if random generated probability $u$ is more than $\beta$. The choice of head has been formally denoted in equation (\ref{eqn:beta}). The attention probability and weighted representations(or attention head's output) is common for all models as shown in equation (\ref{eqn:modified_attention_output}). However, the attention mechanism of transformer-XL has been further decomposed into different components, so, its detailed implementation has been added in the following paragraph.

Each head of transformer-XL consists of $W_q$, $W_{kE}$, $W_v$ and $W_{kR}$ which project the $d$-dimensional query embedding, key embedding, value representation and key representation's positional encoding respectively to dimension $d_h$. The attention score ($A_{ij,\textbf{M}\textbf{N}}$) and attention probability($\alpha_{ij, \textbf{M}\textbf{N}}$) using the relative positional encoding for stochastic cross-head attention for the $\mathbf{M^{th}}$ head with the $\mathbf{N^{th}}$ head and the output of the $\textbf{M}^{th}$ attention head for the $i^{th}$ token is the weighted average of the representations based on the attention probabilities denoted as $C_{i, \textbf{M}\textbf{N}}$are as follows:
\[
     A_{ij,\textbf{M}\textbf{N}}  =  E_{x_i}^{T} W_{q, \textbf{M}}^{T} W_{kE, \textbf{N}} E_{x_j} +
\]
\[
     E_{x_i}^{T} W_{q, \textbf{M}}^{T} W_{kR, \textbf{N}} \mathbf{R_{i-j}} + 
\]
\begin{equation}
\label{eqn:chattn}
     \mathbf{u}^{T}  W_{kE, \textbf{N}} E_{x_j} + \mathbf{v}^{T} W_{kR, \textbf{N}}\mathbf{R_{i-j}}    
\end{equation}

\[
    \alpha_{ij, \textbf{M}\textbf{N}}= \frac{exp( A_{ij, \textbf{M}\textbf{N}} )}{\sum_{m= i-S}^{i} exp( A_{im, \textbf{M}\textbf{N}} )}
\]

\begin{equation}
\label{eqn:modified_attention_output}
C_{i, \textbf{M}\textbf{N}} = \sum_{m= i-S}^{i} \alpha_{im, \textbf{M}\textbf{N}} (W_{v,\textbf{N}}  x_{m})
\end{equation}

The choice of the heads $M$ and $N$ in each layer is based on a generated random number $u$ as follows:


\[ 
    M = h_m \in (h_1 ,. . ,  h_{N_h}) , \text{ } \forall u
\]

\begin{equation}
\label{eqn:beta}
    N= 
    \begin{cases}
        M ,  \text{ if  } u \geq \beta \\
        h_m \in rndprm(h_1,. . ,  h_{N_h})\}  \text{,else}
    \end{cases}
\end{equation}

where $N_h$ denotes the number of heads in the layer, $h_i$ denotes the $i^{th}$ head in a given sequence ($h_1, h_2,.., h_i, .., h_{N_h})$ and rndprm(.) denotes a random permutation of a sequence. We use this method only at the time of training. We experimented with different values of $\beta$ and found $0.1$ to be its optimal value.

\section{Experiments}
\label{Experiments}

In this section, we present the experimental results of our Skip Cross-Head Transformer-XL and compare them with state-of-art benchmarks.

\textbf{Datasets:} We use two datasets namely WikiText-103 ~\citep{DBLP:conf/iclr/MerityX0S17} and enwik8 ~\citep{mahoney2011large}. WikiText-103 is a word level language modelling dataset containing 103 million tokens. It contains tokens from 28K articles. We report perplexity (PPL) of our model on this dataset for $dev$ as well as $test$ sets.

Enwik8 is the character level language modelling dataset that contains 100 million tokens. Enwik8 contains the character tokens from the processed wikipedia text.  We report bits per character(BPC) for $dev$ as well as $test$ sets for this dataset.

The pre-processing and train, validation and test splits of datasets are done similar to \citet{DBLP:conf/acl/DaiYYCLS19}. We have added the complete implementation details in Appendix \ref{appendix:params}.

\subsection{Main Results}

The comparison of the performance of Skip Cross-Head Transformer-XL with other models on WikiText-103 and enwik8 datasets have been shown in Table \ref{table:wt103} and Table \ref{table:enwik8} respectively. This shows the efficiency of the proposed techniques in comparison to other approaches. More importantly, our proposed methods improve the results on our baseline and we do not think our proposed methods should be directly compared with other models apart from baseline because our methods can be applied to other methods as well.

In Table \ref{table:wt103}, our proposed model outperforms the baseline and other models with almost 20\% fewer parameters.

In Table \ref{table:enwik8}, we notice that our model performs quite close to \citep{DBLP:conf/acl/SukhbaatarGBJ19}. We showed in equation \ref{eqn:expected_context}, the expected context that can be attended by our model. We used memory component of $M=512$ and a 12 layer network for Table \ref{table:enwik8} results, which boils down to following expected context as
$
     E(\widetilde{m})  \approx  M \frac{(N-3)}{2} = 512 \frac{(12-3)}{2} = 2304
$.

However \citep{DBLP:conf/acl/SukhbaatarGBJ19} used a context of size 8192 during training as well as evaluation, which requires much larger computational memory when compared to the expected context that our model can attend to during training with constant computational memory. We use larger memory component only during evaluation and that too is limited to size of 3800 only (Table \ref{table:enwik8}).

\begin{table*}
\centering
\begin{tabular}{lllll}
\hline
\textbf{Model} & \textbf{Context} & \textbf{\#Params} & \textbf{Dev} & \textbf{Test}\\
\hline
 Large mLSTM ~\citep{DBLP:conf/iclr/Krause0RL17} & -- & 46M & -- & 1.24 \\
  12L Transformer ~\citep{Al-Rfou_Choe_Constant_Guo_Jones_2019} & -- & 44M  & --  &  1.11  \\
  64L Transformer ~\citep{Al-Rfou_Choe_Constant_Guo_Jones_2019} & -- & 235M  & --  &  1.06  \\
  Adaptive Attention Span~\citep{DBLP:conf/acl/SukhbaatarGBJ19} & 8192 & 39M  & 1.04  & 1.02 \\

\hline
  Transformer-XL ~\citep{DBLP:conf/acl/DaiYYCLS19} & 3800 & 41M  & --  & 1.06   \\
Skip Cross-Head Transformer-XL (Ours) & 1600 & 41M &  1.060 & 1.037  \\
Skip Cross-Head Transformer-XL ( Ours ) & 3800 & 41M & 1.058 &  1.033  \\

\hline
\end{tabular}
\caption{
$Dev$ and $test$ BPC for our proposed model and other approaches on the enwik8 dataset.
}
\label{table:enwik8}
\end{table*}

\begin{table*}
\centering
\begin{tabular}{lllll}
\hline
\textbf{Model} & \textbf{Context} & \textbf{\#Params} & \textbf{Dev} & \textbf{Test}\\
\hline

 DEQ transformer ~\citep{DBLP:conf/nips/BaiKK19} & -- & 138M & -- & 32.40  \\
 
LSTM + Hebbian +cache +MbPA ~\citep{DBLP:conf/icml/RaeDDL18} & -- & --  & 29.0 & 29.2  \\

FNetAR \citep{DBLP:journals/corr/abs-2107-10932} & -- & 144M & -- & 25.81 \\

Transformer-N \citep{DBLP:conf/naacl/SunI21} & -- & 148M &  24.1 & 25.2 \\

Hybrid H3 \citep{DBLP:journals/corr/abs-2212-14052} & -- & 125M & -- & 23.7 \\

\hline
 Transformer-XL~\citep{DBLP:conf/acl/DaiYYCLS19} & 640 & 151M  & --  & 24.03    \\
Skip Cross-Head Transformer-XL (Ours) & 640 & 122M & 22.88 & 23.92  \\
Skip Cross-Head Transformer-XL (Ours) & 2000 & 122M & 21.87  &  22.91 \\

\hline

\end{tabular}
\caption{
$Dev$ and $test$ perplexities (PPL) for our proposed model and other approaches on the WikiText-103 dataset. Our model gives better performance with fewer parameters.
}
\label{table:wt103}
\end{table*}


\begin{table*}
\centering
\begin{tabular}{c|cccccccccc}
\hline
 & \textbf{H1} & \textbf{H2} & \textbf{H3} & \textbf{H4} & \textbf{H5} & \textbf{H6} & \textbf{H7} & \textbf{H8}  & \textbf{stddev} & \textbf{\% stddev change} \\
\hline
Layer 1 & \textbf{-0.01} & \textbf{-0.01} & \textbf{-0.01} & 0.77  & \textbf{-0.01}  & \textbf{-0.01} & \textbf{-0.01} & \textbf{-0.01}  & 0.28 \\

& 0.00 & \textbf{-0.02} & \textbf{-0.01} & \textbf{-0.01}  & 0.0  & \textbf{-0.01} & \textbf{-0.02} & \textbf{-0.01}  & 0.008 & -97.1 \\

\hline

 Layer 2 & 1.61 & 1.9 & 0.13 & 0.19 & 0.03 & 0.12 & 0.13 & 0.03 & 0.77 \\
  & 0.47 & 0.16 & 0.56 & 0.62 & 0.21 & 0.26 & 0.16 & 1.9 & 0.58  & -24.6 \\
 \hline
 
Layer 3 & 0.12 & 1.31 & 0.11 & 0.25 & 1.20 &  0.13 & 0.49 & 0.13  & 0.50  \\

 & 0.09 & 0.29 & 0.11 & 0.06 & 0.12 & 0.60 & 0.20 & 0.15 & 0.18 & -64 \\
\hline
Layer 4 & 0.92 & 0.08 & 0.17 & 0.41 & 0.63 & 0.30 & 0.05 & 0.17 &0.30   \\

 & 0.15 & 0.55 & 0.16 & 0.09 & 0.13 & 0.08 & 0.21 & 0.18  & 0.15 & -50\\
 \hline
 
 Layer 5 & 0.56 & 0.46 & 0.39 & 0.34 & 0.23 & 0.32 & 0.53 & 0.99 & 0.23 \\
 
  & 0.15 & 0.16 & 0.18 & 0.21 & 0.09 & 0.15 &  0.16 & 0.29 & 0.06 & -73.9\\
\hline

Layer 6 & 0.19 & 0.47 & 0.31 & 0.36 & 0.61 & 0.46 & 0.51 & 0.24 &  0.14 \\
 & 0.17 & 0.17 & 0.15 & 0.07 & 0.31 & 0.17 & 0.27 & 0.19 & 0.07 & -50 \\
\hline

Layer 7 &  0.46 & 0.73 & 0.34 & 0.57  & 0.49 & 0.44 & 0.37 & 0.30  &  0.14 \\
 & 0.27 & 0.23 & 0.13 & 0.20 & 0.28 & 0.15 & 0.23 & 0.16 &  0.06 & 57.1 \\
\hline

Layer 8 &  0.11 & 0.22 & 0.17 & 0.53 & 0.22 & 0.42 & 0.09 & 0.25 & 0.15 \\
 & 0.04 & 0.36 & 0.13 & 0.05 & 0.08 & 0.16 &  0.14 & 0.10 & 0.10 &-33.33\\

\hline
\end{tabular}
\caption{
Change in the Perplexity scores (PPL) after pruning of individual attention heads in different layers of Transformer-XL (shown in upper line of each layer's cell) and Transformer-XL with stochastic cross-head attention ($\beta$ = 0.1 ) ( shown in lower line of each layer's cell) . Bold text shows that pruning that head shows improvement on the perplexity (PPL) on WikiText-103 dataset. $Hi$ denotes ith Head
}
\label{table:prune}
\end{table*}

\begin{table*}
\centering
\begin{tabular}{cccc}
\hline
Task &  \textbf{BERT- base} &  \textbf{BERT- base + SCH attn} & \textbf{\% change in std dev(SCH attn)} \\
\hline
MRPC (Acc.) &  82.242  $ \pm$  1.281 &  \textbf{82.609}  $\pm$ 1.4  & 9.3\\
\hfill (F1) & 87.022  $ \pm $  0.988 &  \textbf{87.177} $\pm$ \textbf{0.91}  & -7.9   \\
COLA & 59.849 $\pm$ 1.23 & 58.744 $\pm$ \textbf{1.1} & -10.6 \\
SST2  &  93.038 $\pm$ 0.333  &  92.842 $\pm$  \textbf{0.192} & -42.3\\
MNLI & 84.25 $\pm$ 0403 &  \textbf{84.39} $\pm$ \textbf{0.186} & -53.8\\
QNLI & 91.658 $\pm$ 0.303 &  91.436 $\pm$ \textbf{0.216} & -28.7\\
QQP (Acc) &  91.394 $\pm$ 0.078 &  \textbf{91.422} $\pm$ \textbf{0.042} & -46.1 \\
\hfill (F1) & 88.428 $\pm$ 0.107 &   \textbf{88.436} $\pm$ \textbf{0.073} & -31.8\\ 

\hline

\end{tabular}
\caption{
Effect of finetuning using stochastic cross-head attention with $\beta$= 0.1 on BERT \citep{DBLP:conf/naacl/DevlinCLT19}. All the scores are on $dev$ dataset except for MRPC results with $test$ scores. (Scores: MRPC- Acc/F1, COLA- Mathews corr, MNLI- Acc, SST-2-Acc, QNLI- Acc, QQP- Acc/F1). Mean and standard deviations of ten runs for MRPC and COLA tasks are reported and that of five runs for remaining tasks.
}
\label{table:bert}
\end{table*}


\begin{table*}
\centering
\begin{tabular}{lccc|cc}
\hline
\textbf{Model} & \textbf{SCH attn}  & \textbf{Dev} & \textbf{Test} & Time/batch & Walltime\\
\hline
 Transformer-XL (Our implementation) & $\times$ & 25.64 & 26.70  & 607  & 94:26 \\
Transformer-XL (Our implementation)  & \checkmark & 25.09  &  26.12  & 608 & 105:27 \\
Transformer-XL + Phase 1 training & $\times$ &  25.20 & 26.28  & 535 &  73:48 \\
Transformer-XL + Phase 1 + Phase 2 training & $\times$ &  24.25 & 25.29  & 535, 607 & 95:21  \\
\hline

Skip Cross-Head Transformer-XL & \checkmark &  24.08  & \textbf{25.08}  & 538, 610 &  105:25  \\

\hline
\end{tabular}
\caption{
Ablation study: $dev$ and $test$ perplexity (PPL) for our proposed model on the WikiText-103 dataset. SCH attn denotes the stochastic cross-head attention with $\beta$= 0.1. Phase 1 of the training is based on eqn (\ref{eqn:functionalskip}). Skip Cross-head transformer-XL uses phase 1 and phase 2 of training along with stochastic cross-head attention. Time per batch is in milliseconds and walltime is in hr:min format. $a,b$ format in the time/batch column shows the $a$ milliseconds and $b$ milliseconds are required for the phase 1 and phase 2 of training.  
}
\label{table:ablations}
\end{table*}

We applied our SCH attention mechanism on BERT(Table \ref{table:bert}) and it shows that distribution of information helps in tasks which require semantic information like MRPC, QQP and MNLI, while other tasks show almost similar performance. Also, it should be noted that distribution of information is not as effective here as BERT is an encoder based model and it already captures most of the information through bidirectional context and quite similar information can be transferred even by attending to different subspaces. However, we still notice our mechanism's regularization effect, because the standard deviation of the scores of different runs on almost all the tasks are lower by large margins(almost lower by 30\% on most GLUE tasks).

Similar to \citep{DBLP:conf/nips/MichelLN19}, we also conducted experiments on pruning of individual heads on the transformer-XL with SCH attention mechanism. We can notice that our mechanism helps in regularizing and redistributing the information among different heads of same layer.

\subsection{Ablation Studies}
\label{Ablation_Study}

\begin{table}[!htb]

\begin{tabular}{lrl}
\hline \textbf{$\beta$} & \textbf{Dev} & \textbf{Test} \\ 
\hline
0.0 (Transformer-XL) & 25.64 & 26.70 \\
\hline
0.04  & 25.53 & 26.56   \\
\textbf{0.1} & 25.09 & \textbf{26.12} \\
0.2 & 25.59  &  26.49 \\
0.5 & 26.63 & 27.57 \\
\hline
\end{tabular}
\caption{ Ablation study on $\beta$ for SCH attn. } 
\label{abl:betawt103} 
\end{table}

\begin{table}[!htb]
\centering
\begin{tabular}{lrl}
\hline \textbf{$p_{skip}(layer)$} & \textbf{Dev} & \textbf{Test} \\ \hline
Equation (\ref{eqn:uniformskip}), p= 0.1 & 25.63 & 26.66  \\
Equation (\ref{eqn:uniformskip_initial}), p= 0.1 & 25.20 & 26.26  \\
Equation (\ref{eqn:uniformskip_final}), p= 0.1 & 25.66 & 26.56 \\
Equation (\ref{eqn:uniformskip_both}), p= 0.1 & 25.15 & 26.14 \\

\hline
\end{tabular}
\caption{\label{font-table} Ablation study on the effect of skipping first layer, final layer and both first and final layers of the model. Results are shown after Phase 1 (Skiip-Retain training phase) of our training mechanism.}
\label{table:skipwayswt103}

\end{table}


\begin{table}[!htb]

\centering
\begin{tabular}{lrl}
\hline \textbf{$p_{skip}(layer)$} & \textbf{Dev} & \textbf{Test} \\ \hline
Eqn (\ref{eqn:uniformskip}), p= 0.1 & 25.63 & 26.66  \\
Eqn (\ref{eqn:uniformskip}), p= 0.3 & 28.56 & 29.63  \\
Eqn (\ref{eqn:uniformskip_both}), p= 0.1 & 25,15 & 26.14  \\
Eqn (\ref{eqn:functionalskip}) &  25.20 & 26.28  \\

\hline
\end{tabular}
\caption{ Ablation study for Phase 1 (Skip-Retain training phase) of our training mechanism.}
\label{table:sdwt103}
\end{table}

\begin{table}[!htb]
\centering

\begin{tabular}{lrl}
\hline \textbf{$p_{skip}(layer)$} & \textbf{Dev} & \textbf{Test} \\ \hline
Transformer-XL &  25.64 & 26.70 \\
\hline
Eqn (\ref{eqn:uniformskip_both}), p= 0.1 & 24.55 & 25.47  \\
Eqn (\ref{eqn:functionalskip}) &  24.25 & \textbf{25.29}  \\

\hline
\end{tabular}
\caption{ Ablation results after Phase 1 and Phase 2 (vanilla training phase) of our training mechanism.}
\label{table:sdfinewt103}

\end{table}


We have shown an overview of the final ablation results in Table \ref{table:ablations}. Our detailed ablation experiments are described in the following paragraphs.

In Table \ref{abl:betawt103}, We found $0.1$ to be the optimal value for stochastic cross-head attention probability showing that with $0.1$ probability most of the information can be distributed among heads of same layer and thereby, helping as an internal regularizer.

In Table \ref{table:skipwayswt103}, we experimented with uniformly skipping initial or final layers or both layers where we keep the probability $p$ as constant for fair comparison. We show that the initial layer should not be skipped as it captures local dependencies among tokens and final layer should not be skipped because it produces logits that are used for calculation of categorical probability over next token.

In Table \ref{table:sdwt103}, we experimented to find out whether a linear skipping of layers(equation \ref{eqn:functionalskip}) is optimal or uniformly skipping the layers with same probability (based on equation (\ref{eqn:uniformskip_both})) is optimal. We experimented with different values of $p$ for uniform skipping.
We found that using eqn (\ref{eqn:functionalskip}) to be optimal (Table \ref{table:sdwt103} and Table \ref{table:sdfinewt103} ). Eqn (\ref{eqn:functionalskip}) is optimal because similar to CNNs, transformer layers attend to a local context in its initial layers and upper layers attend to much global context and thus, the upper layers need a longer context to produce better results. However, using a uniform skipping in every layer, we can only attend to similar past context in every layer.

\section{Conclusions}

Our Skip-Retain training mechanism improves deep representations in memory-based transformers without any additional memory requirements. Our regularization mechanism does not require any additional parameters and can be used a plug and play mechanism for other transformer models as well.



\section*{Acknowledgements}

We thank HPCE, IIT Madras that provided resources to carry out the experiments. We also thank the faculties, family and friends whose support and help made this work possible.

\bibliography{custom}
\bibliographystyle{acl_natbib}

\appendix

\section{Recap: Transformer-XL}
\label{appendix:recap_txl}

Transformer-XL ~\citep{DBLP:conf/acl/DaiYYCLS19} is based on transformer architecture. Each layer of  architecture additionally consists of a memory component which stores the past input representations given to that layer. The memory component does not allow gradients to pass through these past representations. Transformer-XL also introduced the relative positional encoding scheme.


Each layer gets as input a set of token representations $ \{ x_{t+1}, x_{t+2}, .. ,  x_{t+L} \} $, where $L$ is the segment length (also called block size). Each layer stores a set of past activations $ \{ x_{t-M+1}, x_{t-M+2}, x_{t- M+3}, ..,  x_{t} \} $ in the memory component where $M$  denotes the number of past token representations stored. Each token is represented with a $d$-dimensional representation.

\textbf{Multi-head attention module} : This is the core component of transformer layer. It consists of multiple attention heads that work in parallel and transforms the inputs into query, key and value representations, having the dimensions $d_h$, using the trainable projection matrices denoted as $W_q$, $W_k$ and $W_v$  $\in \mathbb{R}^{d_h \times d}$.

The query representation of $x_{i}$ finds the similarity with the key representaions of \{$x_{i-S+1}, x_{i-S+2},. . ,   x_{i}$ \} using a dot product. Here $S$ is the context length and S $\in$[$i-M+1,i$]. The attention score ($A_{ij}$) between the query representation of $x_{i}$ token and the key representation of $x_j$ token is calculated as follows:
\begin{equation}
\label{eqn:score}
     A_{ij}= x_{i}^T  W_q^T  W_k x_{j}
\end{equation}

The attention probabilities ($\alpha_{ij}$) are obtained by using softmax on these attention scores as follows:
\begin{equation}
\label{eqn:attn_prob}
    \alpha_{ij}= \frac{exp( A_{ij} )}{\sum_{m= i-S}^{i} exp( A_{im} )}
\end{equation}

The output of the attention head for the $i^{th}$ input token (denoted as $C_i$) is obtained by using the weighted average of the value representations based on the attention probabilities:
\begin{equation}
\label{eqn:attn_output}
    C_{i} = \sum_{m= i-S}^{i} \alpha_{im}(W_v x_{m})
\end{equation}
The outputs obtained from different heads are concatenated and then multiplied with a trainable output matrix (denoted as $W_o$)  to obtain the final output for the mutli-head attention module.

\textbf{Feedforward network module} :  This module is a feedforward network containing a linear layer followed by a non linear layer and a linear layer.

\textbf{Relative positional encoding scheme}: The token representations in (\ref{eqn:score}) consist of the embedding representations ($E$) and the positional encoding representations ($U$) of the input tokens. Thus, tokens $x_i$ and $x_j$ can be represented as $x_i$= $E_{x_i} + U_{x_i}$ and $x_j$= $E_{x_j} + U_{x_j}$ respectively. Equation (\ref{eqn:score}) can be expanded as follows:
\[
     A_{ij}  =  (E_{x_i} + U_{x_i})^{T} W_q^{T} W_k (E_{x_j} + U_{x_j})
\]
\[
     A_{ij}  =  E_{x_i}^{T} W_q^{T} W_k E_{x_j} + E_{x_i}^{T} W_q^{T} W_k U_{x_j} + 
\]
\begin{equation}
\label{eqn:score_expand}
 U_{x_i}^{T}  W_q^{T}  W_k E_{x_j} +
 U_{x_i}^{T}  W_q^{T}  W_k U_{x_j}    
\end{equation}

The learnable parameters in (\ref{eqn:score_expand}) are as follows:

\[
     A_{ij}  = \underbrace{ E_{x_i}^{T} W_q^{T} W_{k E} E_{x_j} }_\text{(A)} + \underbrace{E_{x_i}^{T} W_q^{T} W_{k R} \mathbf{R_{i-j}} }_\text{(B)} + 
\]

\begin{equation}
\label{eqn:relpos}
\underbrace{ \mathbf{u}^{T}  W_{k E} E_{x_j} }_\text{(C)} + \underbrace{ \mathbf{v}^{T} W_{k R}\mathbf{R_{i-j}} }_\text{(D)}
\end{equation}

where $\mathbf{R_{i-j}}$ denotes the relative positions and $\mathbf{u}$ and $\mathbf{v}$  $ \in \mathbb{R}^{d_h}$ are global learnable parameters. The $W_q$, $W_k$, $W_{kE}$ and $W_{kR}$ $\in \mathbb{R}^{d_h \times d}$ are the trainable weight matrices. In equation (\ref{eqn:relpos}), Part (A) is the content based addressingterm, Part (B) is the content dependent positional bias term, Part (C) is the global content bias term and Part (D) is the global positional bias term.

The relative position ($r$) is a function that maps the $i^{th}$ token in an input sequence \{$x_1, x_2,.., x_l$\} to its relative position with respect to the last token in the sequence, i.e. $r(x_l)$ = 0, $r(x_{l-1})$ = 1, .., $r(x_2)$= $l-2$ and $r(x_1)$ = $l-1$.


Positonal Encoding(PE) of a token in a sequence for the $i^{th}$ dimention is calculated as follows:
\begin{equation}
\label{eqn:PE}
    PE_{r, i}= 
    \begin{cases}
        sin (r/ 10000^{2i/ d}) , \text{ if }  0 \leq i \text{ } < \text{ } \frac{d}{2} \\
        cos (r/ 10000^{2(i- d/2)/ d}) , \text{ if } \frac{d}{2} \leq i < d
    \end{cases}
\end{equation}

The relative positional encoding ($\mathbf{R_{i-j}}$ in (\ref{eqn:relpos})) of a token with relative posittion $r$ is a $d$-dimensional representation calculated as follows:
\begin{equation}
\label{eqn:relPE}
    \mathbf{R_{i-j}} = concat(PE_{r, 0}, PE_{r, 1}, . . , PE_{r, d-1}) 
\end{equation}

\section{Model Hyperparameters and
experiment settings}
\label{appendix:params}

\subsection{Ablation study settings}
For our ablation studies, we used batch size of 50 with block size (or segment length $L$) of 512 tokens (i.e. effective batch size of 25.6K tokens). A network with 8 layers was used for the ablation studies. The hidden size ($d$) of 512 and hidden size in the middle layer of feedforward module (inner hidden size or $d_{inner}$) of 2048 was used. Memory component length ($M$) of 512 was used. The number of heads used in each layer was 8 and the head dimension($d_h$) was 64. We used a dropout of 0.1 and attention dropout and embedding dropout to be 0. We used adaptive embedding and softmax implementations as \citet{DBLP:conf/iclr/BaevskiA19} and \citet{DBLP:conf/icml/GraveJCGJ17} with adaptive cutoffs as [20000,40000, 200000] and adaptive division value as 4. Learning rate of 0.00025 was used with Adam ~\citep{DBLP:journals/corr/KingmaB14} optimizer. Cosine annealing learning rate scheduling ~\citep{DBLP:conf/iclr/LoshchilovH17} was used  with max iterations as 16000. For our ablation studies, we stopped the training when the change in perplexity score was less than 0.2 for 64K training steps. During evaluation,context length of 640 was used.

For the ablation study on the effect of using stochastic cross-head attention on number of heads in the transformer-XL (Table \ref{table:headwt103}), we used the hidden size ($d$) as 480 and inner hidden size ($d_{inner}$) as 1920. Remaining hyperparameters were the same as in the other ablation studies.

At the time of evaluation, we used a batch size of 10 with block size of 128 (effective batch size of 1280 tokens). Also, at the time of evaluation the stochastic cross-head attention mechanism is not used. Remaining hyperparameters remain the same while evaluation.

\subsection{Main Results settings}
For the enwik8 model(Table \ref{table:enwik8}), batch size of 60 with block size (or segment lenth ($L$) of 512 tokens  was used. A network with 12 layers was used. The hidden size($d$) of 512 and hidden size in the middle layer of feedforward module (inner hidden size or $d_{inner}$) of 2048 was used. The number of heads used in each layer was 8 and the head dimension ($d_h$) was 64. Cosine annealing learning rate scheduling ~\citep{DBLP:conf/iclr/LoshchilovH17} was used  with max iterations as 4000.


For the WikiText-103 model(Table \ref{table:wt103}), batch size of 32 with segment length ($L$) of 512 tokens was used. A network with 15 layers was used. The hidden size ($d$) of 624 and inner hidden size ($d_{inner}$) of 2496 was used.  16 heads were used with head dimension ($d_h$) of 39. All the remaining parameters remain same as used in \nameref{Ablation_Study}. 

For both the above mentioned models, We used a dropout of 0.15 and attention dropout and embedding dropout to be 0. Memory component length ($M$) of 512 was used. Learning rate of 0.00025 was used with Adam ~\citep{DBLP:journals/corr/KingmaB14} optimizer.

For the BERT experiments in Table \ref{table:bert}, We used $\beta = 0.1$. Remaining hyperparameters were same as \citet{DBLP:conf/naacl/DevlinCLT19}. We used stochastic-cross head attention only for finetuning on the GLUE tasks.

The evaluations of all the models were carried out as described in \nameref{Ablation_Study}. All of the experiments were conducted on two Nvidia Tesla V100 GPUs with memory capacity of 32 GB each.

\subsection{Stochastic Cross-Head Attention Mechanism: Experiments with number of attention heads}

Table \ref{table:headwt103} shows the effect of using stochastic cross-head attention with Transformer-XL. Here, we show that performance is improved to a large extent by using this mechanism and increasing the number of heads at the same time. However, this needs further experiments to confirm that it improves the performance with an increase in the number of heads.

\begin{table}
\centering
\begin{tabular}{ccc}
\hline \textbf{\#Heads} & \textbf{TXL} & \textbf{SCH attn} \\ \hline
4  & 27.53 & 26.82\\

8 & 26.99 &  26.50 \\
12  &  27.40 & 26.76 \\
16 &  27.42 & 26.41 \\
\hline
\end{tabular}
\caption{ The effect of our attention mechanism (SCH attn) with $\beta=0.1$ on the number of heads of transformer-XL model.}
\label{table:headwt103}
\end{table}

\end{document}